\documentclass{article}

\usepackage{arxiv}
\usepackage{multirow} 
\usepackage[utf8]{inputenc}
\usepackage[T1]{fontenc}
\usepackage{times}
\usepackage{helvet}
\usepackage{courier}

\usepackage{amsmath,amssymb,amsfonts}
\usepackage{graphicx}
\usepackage{booktabs}
\usepackage{nicefrac}
\usepackage{microtype}
\usepackage{caption}
\DeclareCaptionType{listing}[Listing][List of Listings]
\usepackage{natbib}
\usepackage{algorithm}
\usepackage{algorithmic}

\usepackage[many]{tcolorbox}
\tcbuselibrary{listings, skins, breakable}

\newtcolorbox[auto counter, number within=section]{listingbox}[2][]{%
  enhanced,
  breakable,
  colback=gray!5,       % 背景色 (浅灰)
  colframe=black!30,    % 边框色 (深灰)
  coltitle=black,       % 标题文字颜色
  fonttitle=\bfseries,
  title=Listing~\thetcbcounter:~#2, % 自动编号 + 标题
  sharp corners,
  boxrule=0.8pt,
  left=6pt, right=6pt, top=6pt, bottom=6pt,
  before skip=10pt, after skip=10pt,
  #1
}

\PassOptionsToPackage{hyphens}{url}
\usepackage{xcolor} 
\usepackage[colorlinks=true, linkcolor=blue, citecolor=blue, urlcolor=blue]{hyperref}

\title{Benchmarking Multi-Step Legal Reasoning and Analyzing Chain-of-Thought Effects in Large Language Models}

\author{
Wenhan Yu\textsuperscript{1}\thanks{Equal contribution.} \quad
Xinbo Lin\textsuperscript{2}\footnotemark[1] \quad
Lanxin Ni\textsuperscript{3} \quad
Jinhua Cheng\textsuperscript{2}\thanks{Corresponding authors.} \quad
Lei Sha\textsuperscript{1}\footnotemark[2] \\
\textsuperscript{1}School of Artificial Intelligence, Beihang University, Beijing, China \\
\textsuperscript{2}KoGuan School of Law, Shanghai Jiao Tong University, Shanghai, China \\
\textsuperscript{3}School of Criminal Law, China University of Political Science and Law, Beijing, China \\
\texttt{\{yuwenhan, shalei\}@buaa.edu.cn},\, \\
\texttt{\{linxinbo, chengjinhua\}@sjtu.edu.cn},\, \\
\texttt{lanxin.ni.law@gmail.com}
}

\begin{document}

\maketitle

\begin{abstract}
As large language models (LLMs) exhibit advanced reasoning capabilities in different specialized domains, their application to legal reasoning tasks is actively being explored. The pursuit of justice in legal contexts demands not only correct outcomes but also \textbf{reasoned elaboration}, which necessitates deriving conclusions through logically justified and transparent argumentation. Current legal benchmarks, despite being cited as reasoning-focused, suffer from three critical limitations: conflation of factual recall with genuine inference, fragmentation of holistic reasoning processes, and neglect of reasoning process quality. To bridge these gaps, we construct \textbf{MSLR, the first Chinese multi-step legal reasoning dataset} centered on legal decision-making. To align with real-world legal reasoning trajectories, MSLR employs the \textbf{IRAC framework} (Issue-Rule-Application-Conclusion) to capture expert reasoning traces from official legal decisions. In parallel, we design a scalable \textbf{Human-LLM collaborative annotation pipeline} that efficiently generates fine-grained step-level annotations while establishing a reusable methodological framework for multi-step reasoning datasets. Evaluation of a range of LLMs on MSLR reveals only modest performance, highlighting substantial challenges in adapting to {complex legal reasoning}. Further experiments show that \textbf{Self-Initiated CoT prompts}—created autonomously by the models—consistently improve reasoning coherence and output quality, outperforming Human-Designed CoT prompts, which often yield ambiguous results. This work contributes to the broader discourse on LLM reasoning and CoT strategies, offering practical insights and resources for future research. \textbf{The dataset and code are publicly available at} \href{https://github.com/yuwenhan07/MSLR-Bench}{https://github.com/yuwenhan07/MSLR-Bench} \textbf{and} \href{https://law.sjtu.edu.cn/flszyjzx/index.html}{https://law.sjtu.edu.cn/flszyjzx/index.html}.
\end{abstract}

%%%%%%%%%%%%%%%%%

\begin{figure*}[t]
\centering
\includegraphics[width=\textwidth]{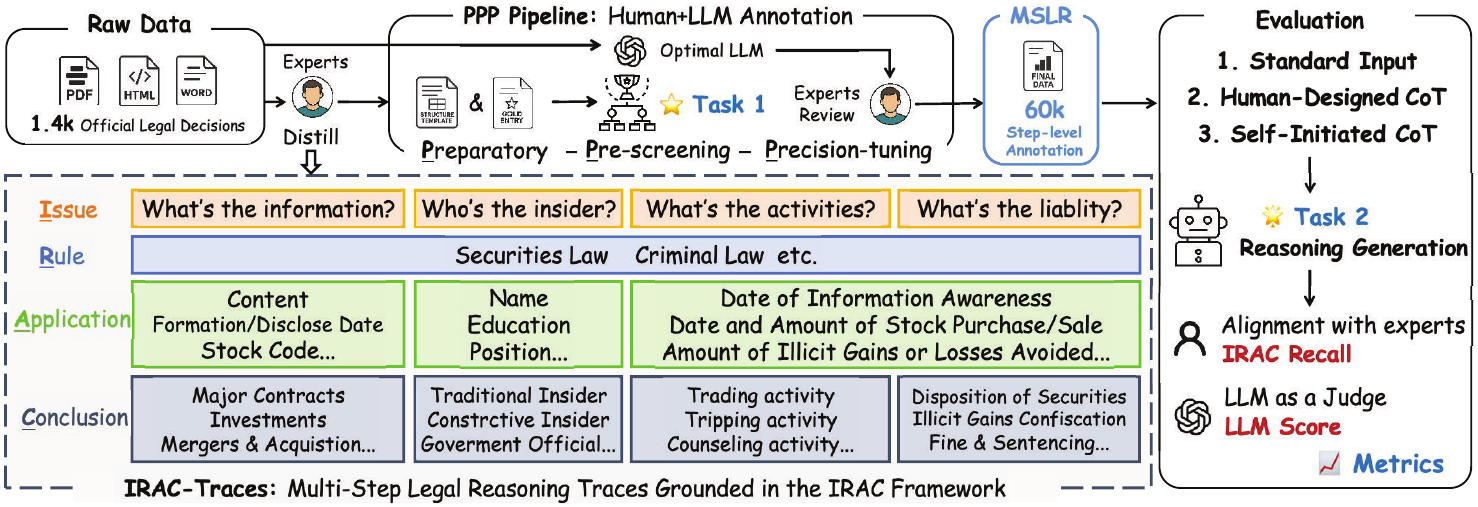}% Reduce the figure size so that it is slightly narrower than the column.
\caption{A schematic of the MSLR construction and evaluation framework.}
\label{figure1}
\end{figure*}

\section{Introduction}
Recent breakthroughs in the reasoning capabilities of large language models (LLMs) \cite{Huang_Chang_2023, openai2024openaio1card, deepseekr1} have spurred increasing interest in leveraging them for legal reasoning tasks \cite{NLP_legal_survey}. As LLMs are applied to various legal domains, including legal document summarization \cite{ashTranslatingLegaleseEnhancing2024a, liuLowresourceCourtJudgment2024a, moroMultilanguageTransferLearning2024a}, legal question answering \cite{AnsweringLegalQuestions2024, InterpretableLongformLegal2024a, DiscoLQA2025}, and legal judgment prediction \cite{dengSyllogisticReasoningLegal2023, tongLegalJudgmentPrediction2024b, wangLegalReasonerMultiStageFramework2024}, evaluating their capabilities and limitations becomes critically important. To address this need, specialized legal benchmarks (e.g., LexGLUE, \textsc{LegalBench}, LEXTREME, LawBench, LexEval) have emerged. These benchmarks provide targeted datasets and tasks covering legal reasoning, advancing the development of legal artificial intelligence to some extent.

However, these benchmarks exhibit three core limitations. First, they often conflate factual recall with genuine legal reasoning. Tasks range from retrieval-based (e.g., \textit{“What is Article 180 of the Criminal Law?”}) to reasoning-intensive (e.g., \textit{“Is the defendant liable?”}), making it hard to isolate reasoning capabilities.

Second, benchmarks such as \textsc{LegalBench} decompose multi-step reasoning (e.g., \textbf{IRAC} \cite{IRAC}) into disconnected subtasks—issue spotting, rule recall, application, and conclusion—assessed separately on heterogeneous data. This fragmented design fails to capture the coherent, step-by-step nature of real-world legal reasoning.

Third, current evaluations prioritize final answers while ignoring the reasoning process itself. Yet legal decisions require not just correct outcomes but also defensible, well-justified logic—known as \textit{Reasoned Elaboration} \cite{ReasonedElaboration1973}. Existing datasets lack annotated reasoning traces, making it impossible to assess \textbf{faithfulness or explainability} \cite{ExplanationAILaw2020}.

To address these gaps, we introduce \textbf{MSLR}, a multi-step legal reasoning benchmark aligned with the IRAC framework (Figure \ref{figure1}). We focus on insider trading—a domain demanding structured legal justification—and collaborate with legal experts to construct standardized \textbf{IRAC-Traces}, annotating 60K step-level labels across 1,400 official legal decisions. This granular stepwise dataset: (1) establishes the foundation for process-quality-oriented evaluation metrics, (2) paves the way for supervising LLMs’ reasoning.

To efficiently scale high-quality annotations, we propose a three-stage Human-LLM collaborative pipeline (\textbf{PPP}): (1) \textbf{P}reparatory stage. Domain experts design annotation templates and criteria, and provide gold-standard examples. (2) \textbf{P}re-screening stage. Multiple LLMs perform preliminary annotation; the optimal model is selected based on quality-cost efficiency against gold-standard examples. (3) \textbf{P}recision-tuning stage. The optimal LLM generates comprehensive annotations refined by human annotators adhering to expert-defined criteria. This pipeline not only maintains annotation quality while significantly reducing marginal costs but also offers methodological insights for constructing multi-step reasoning datasets in other domains.

Using our novel \textit{IRAC Recall} metric and an LLM-as-a-Judge framework \cite{LLMasaJudge2025}, we evaluate general, legal, and reasoning-focused models. Results show that: (1) MSLR is a challenging, unsaturated benchmark (e.g., OpenAI o1-mini achieves only 72\% IRAC Recall); (2) IRAC Recall strongly correlates with expert-aligned LLM judgments; (3) Human-Designed Chain-of-Thought (CoT) prompts can yield diminishing returns, while \textit{Self-Initiated CoT}—generated by the model itself—consistently enhances reasoning quality.

In summary, our main contributions are as follows:
\begin{itemize}
\item \textbf{MSLR Dataset:} We construct the first Chinese multi-step legal reasoning benchmark aligned with the \emph{IRAC} framework, enabling granular evaluation of the quality and coherence of LLM reasoning across nearly 1,400 cases with 60K step-level annotations.
\item \textbf{PPP Annotation Pipeline:} We propose a scalable Human-LLM collaborative pipeline that reduces annotation costs while preserving quality, providing a methodological blueprint for capturing real-world reasoning traces.
\item \textbf{Reasoning-quality-oriented Metrics:} We design dual evaluation metrics demonstrating statistically significant consistency: (1) \emph{IRAC Recall} measuring alignment with expert reasoning traces; (2) \emph{LLM Score} assessing legal logical coherence.
\item \textbf{CoT Prompt Insights:} We reveal the ambiguous outcomes of \emph{Human-Designed CoT} and demonstrate the potential of \emph{Self-Initiated CoT} as a paradigm for high-quality legal reasoning.
\end{itemize}

\section{Related Work}
\subsection{Legal Reasoning Benchmark}
 To foster reliable legal AI applications, multiple international benchmarks for legal reasoning have been established. LexGLUE \cite{LexGLUE2022} and LegalBench \cite{LegalBench2023} encompass diverse English-language legal tasks, LawBench \cite{LawBench2024} and LexEval \cite{LexEval2024} integrate comprehensive Chinese legal domains, while LEXTREME \cite{LEXTREME2023} covers legal tasks across 24 official EU languages. However, most benchmarks oversimplify downstream legal reasoning tasks into single-label or multi-label classification, leading to a narrow focus on end-task accuracy while neglecting the quality of reasoning processes \cite{Huang_Chang_2023}. Consequently, it remains unverifiable whether LLMs can genuinely simulate human-like reasoning processes. Recent studies have developed specialized datasets for complex tasks—such as long-form legal QA \cite{InterpretableLongformLegal2024a}, criminal fact determination \cite{shenLawReasoningBenchmark2025}, and open-ended legal exams \cite{LEXam2025}—to evaluate the quality of LLMs’ reasoning processes. Our work bridges this critical gap by establishing the first benchmark for assessing multi-step reasoning performance in legal decision-making scenarios.

Although human annotation remains indispensable for the construction of high-quality datasets, its high cost and limited scalability present significant barriers. Recent studies propose using LLMs for automated annotation as a cost-effective alternative \cite{Annotation_Survey2024}, yet they face challenges in validation \cite{ChatGPTLabel2024}, particularly in step-by-step reasoning contexts \cite{LetsVerifyStep2023}. A promising paradigm is Human-LLM collaborative annotation \cite{CoAnnotating2023, MEGAnno2024}, which leverages LLMs to accelerate the process while maintaining quality through human validation \cite{HumanLLMCollaborativeAnnotation2024}. Our work advances this paradigm by establishing a scalable annotation pipeline for reasoning trace datasets.

Current legal benchmarks mostly rely on outcome-oriented metrics such as macro-$F_1$, accuracy, or ROUGE, which fail to effectively evaluate the quality and coherence of reasoning processes. Recently, the \textit{LLM-as-a-Judge} \cite{LLMasaJudge2025} has emerged as a common paradigm to assess the plausibility of generated content, but whether LLM is capable of judging the quality of legal reasoning remains an open question. We apply this paradigm to evaluate the quality of legal reasoning and verify its reliability. 

\subsection{Chain-of-Thought}
Chain-of-Thought is a widely adopted prompting technique for LLMs that guides models to think step-by-step \cite{nyeShowYourWork2021, weiChainofThoughtPromptingElicits2022}. Given its extensive study in mathematical reasoning and consistent positive performance \cite{LetsVerifyStep2023}, CoT is commonly assumed to benefit nearly all reasoning tasks. Diverse CoT prompting strategies have been proposed and applied broadly \cite{SurveyPrompt2025}, including automatic \cite{AutomaticCot2022}, self-consistency \cite{SelfConsistencyCot2023}, and logical \cite{LogicCoT2024} CoT. However, a recent systematic meta-analysis reveals that CoT yields significant gains primarily in mathematical and symbolic reasoning tasks, while showing minimal improvements or even detrimental effects in non-symbolic domains such as knowledge, soft reasoning, and commonsense tasks \cite{CoTNotCoT2025}. Since the efficacy of CoT in legal reasoning remains unclear, our work conducts an exploratory analysis to address this gap.

\section{MSLR Dataset}
\subsection{Data Sources}
Our work aims to construct \textbf{a high-density, high-quality multi-step} legal reasoning dataset: \textbf{MSLR}. We therefore focus on legal decisions from Chinese insider trading cases, primarily because these cases exhibit high complexity and logical rigor in both factual details and legal rules. On the one hand, insider trading cases typically involve extensive financial data, temporal information, behavioral patterns and intentions of multiple parties, and long causal chains. On the other hand, their regulatory frameworks encompass securities law, criminal law, numerous judicial interpretations, and regulatory provisions. This duality better reflects the robust information integration and analytical capabilities required of real-world legal professionals in making legal decisions.

Primary data collection encompasses \textbf{1,389} legal documents issued by the China Securities Regulatory Commission (CSRC), the Supreme People's Procuratorate (SPP), and the Supreme People's Court (SPC) between 2005 and 2024. With an average length of \textbf{2,516} Chinese characters per document, this corpus underscores the inherent complexity of legal reasoning processes.

\subsection{Dataset Creation}
This dataset is created and validated by a legal expert team from the Center for Empirical Legal Studies of Shanghai Jiao Tong University, including one professor, two Ph.D. candidates, and four master’s students. To authentically reconstruct holistic legal reasoning processes in real-word, the professor and Ph.D. candidates distill standardized step-by-step reasoning traces based on the IRAC framework: \textbf{IRAC-Traces}. Given the labor-intensive nature of dataset construction, we design a scalable Human-LLM collaborative annotation pipeline structured into three stages:

\textbf{Preparatory stage.} First, based on these IRAC-Traces, we develop structured data annotation templates and criteria to systematically record the reasoning process of each legal decision. Second, graduate students create gold-standard examples. To ensure annotation quality, graduate students undergo a two-week pre-training program to strengthen their legal knowledge and familiarity with annotation guidelines. Before formal annotation, they must perform trial annotations on 50 cases, achieving 95\% alignment with professor annotations. During formal annotation, we adopt a ``2+1" protocol: two annotators independently annotation the same case, and a third annotator reviews and adjudicates discrepancies. This process yields gold-standard annotations for over 900 cases with high inter-annotator agreement (Krippendorff's $\alpha = $ 0.832). 

\textbf{Pre-screening stage.} First, we transform the annotation tasks into structured information extraction tasks suitable for LLMs. Next, we evaluate different LLMs' performance against gold-standard examples to identify the most cost-effective model. Experimental results (see Section ``Results and Analysis'') demonstrate DeepSeek-V3's optimal cost-effectiveness ratio, establishing it as our primary annotation assistant.

\textbf{Precision-tuning stage.} First, building on the optimal model, we further refine prompts for a subset of fields that are semantically dependent, ambiguous, or irregularly formatted. By optimizing prompt templates and conducting independent extraction trials, we significantly enhance the accuracy of information extraction for these challenging fields. Second, to ensure the reliability and accuracy of the optimal model’s annotations, each document’s structured data undergoes item-by-item review by a legal expert. In cases of significant discrepancies, a second legal expert is invited to conduct a joint review, ensuring consensus and precision in the final annotations.

\begin{table}[htbp]
  \centering
  \begin{tabular}{p{0.7\linewidth}r}
    \toprule
    \textbf{Characteristic} & \textbf{MSLR} \\
    \midrule
    Document time span & 2005–2024 \\
    Total number of documents & 1,389 \\
    Avg. characters per document (chars) & 2,516 \\
    Avg. structured fields per document & 43.03 \\
    Total number of structured entries & 59,771 \\
    Avg. fill rate of key fields (\%) & 90.58 \\
    \bottomrule
  \end{tabular}
  \caption{Descriptive statistics of the MSLR dataset.}
  \label{tab:dataset-statistics}
\end{table}

Leveraging the PPP pipeline, we ultimately annotate \textbf{59,771} step-level reasoning labels across 1,389 official legal decisions. This corresponds to an average of 43 intermediate steps per legal decision, demonstrating the dataset’s faithful representation of multi-step reasoning in complex legal contexts. These statistics confirm that the dataset achieves representative scale, structural consistency, and systematic organization. Its stable annotation quality establishes a robust foundation for subsequent experimental evaluation tasks.

To validate the novelty of the MSLR dataset, we compute ROUGE-1/2/L $F_1$ scores between documents in MSLR and the comprehensive Chinese legal corpus used during the training of Lawyer LLaMA \cite{huang2023lawyer}. The results show near-random similarity levels (all \(\ F_1 < 0.03 \)), significantly below conventional similarity thresholds. This indicates that MSLR was not included in the pretraining corpus, further establishing its reliability as a contamination-free evaluation benchmark.

\section{Experiment Setup}

\subsection{Baseline Models}
We evaluate a diverse set of LLMs, grouped into three major categories. \textbf{General models} are primarily designed for instruction following and open-domain tasks. This group includes Llama3-8B-Chinese and Llama3.3-70B-Instruct~\cite{grattafiori2024llama3herdmodels}, GLM4-9B~\cite{glm2024chatglm}, Qwen2.5-7B and Qwen2.5-72B~\cite{qwen2.5}, DeepSeek-V3~\cite{deepseekv3}, GPT-4~\cite{openai2024gpt4technicalreport} and GPT-4o~\cite{openai2024gpt4ocard}.
\textbf{Legal-domain models} are fine-tuned from general models using legal texts such as statutes and court rulings to improve performance in legal contexts. Representative models in this category include AIE-51-8-Law, based on Qwen-2.5-3B Instruct~\cite{qwen2.5}; CleverLaw, based on InternLM3-8B-Instruct~\cite{cai2024internlm2}; Law Justice, based on Llama-3.1-8B Instruct~\cite{grattafiori2024llama3herdmodels}; and Lawyer-LLM~\cite{huang2023lawyer}.
\textbf{Reasoning models} incorporate slow-thinking mechanisms such as Chain-of-Thought, Reflection, or Iterative Verification. This category comprises models such as DeepSeek-R1-Distill-Qwen-1.5B, DeepSeek-R1-Distill-Qwen-7B, DeepSeek-R1-Distill-Llama-70B~\cite{deepseekr1}, QwQ-32B~\cite{qwq32b}, DeepSeek-R1~\cite{deepseekr1}, o1-mini and o3-mini~\cite{openai2024openaio1card}.

All models are evaluated on MSLR dataset, focusing on the two tasks below. To balance generation stability and output diversity, we set the decoding parameters for all models supporting customization to a uniform configuration of \texttt{top-p=0.9} and \texttt{temperature=0.2}. Furthermore, due to computational resource constraints and the cost associated with API usage, we restrict all models to single-pass inference to generate final outputs. No advanced inference techniques, such as ensemble methods or self-consistency sampling, are employed.

\subsection{Task 1: LLM Automatic Annotation}
\subsubsection{Experimental Design}

To evaluate the annotation capabilities of LLMs in the legal domain, we transform annotation tasks into structured information extraction tasks compatible with LLMs. These encompass core NLP sub-tasks including named entity recognition, text classification, and summarization. We design a unified structured information extraction framework that leverages a zero-shot setup to construct standardized prompts, instructing LLMs to extract key information from legal texts and output results in a predefined JSON structure with specified fields.

\subsubsection{Evaluation Metrics}
\label{task1 evaluation}
Two metrics are designed to evaluate the annotation quality of LLMs:

\textbf{(1) Field Completeness Rate (FCR)}, which measures the coverage of model outputs over the fields that should be filled according to the gold-standard examples. Let $N$ be the total number of cases. For the $i$-th case, let $R_i=$ the number of non-empty fields in the reference answers, $P_i=$ the number of fields predicted by the model, and $O_i=$ the number of overlapping fields between prediction and reference. FCR is defined as:

\begin{equation}
\textit{FCR} = \frac{1}{N} \sum_{i=1}^N \frac{O_i}{R_i}.
\end{equation}

\textbf{(2) Overall Accuracy}, which reflects the annotation accuracy of the model in the overlapping fields. We classify fields into two categories: \textbf{Exa-fields} requiring exact matching (e.g.,\textit{ dates}, \textit{amounts}) and \textbf{Sem-fields} requiring semantic matching (e.g., \textit{event descriptions}, \textit{information summaries}). For the $i$-th case, let $Oe_i=$ the number of Exa-fields within overlapping fields, $Os_i=$ the number of Sem-fields within overlapping fields. 

For Exa-fields, \textbf{Exact Accuracy} is used to measure exact matches between predicted values $\hat{y}$ and references $y$ across the $Oe_i$ overlapping fields:
\begin{equation}
\textit{ExaAcc}_i  = \frac{1}{Oe_i} \sum_{j=1}^{Oe_i} \mathbb{I}(\hat{y}_j = y_j),
\end{equation}

where $\mathbb{I}(\cdot)$ is the indicator function. 

For Sem-fields, \textbf{Semantic Accuracy} uses cosine similarity between embeddings $f(\cdot)$ from ChatLaw-Text2Vec~\cite{ChatLaw} to account for semantically equivalent but lexically diverse expressions, considering predictions correct if similarity exceeds threshold $\tau=0.6$:

\begin{equation}
\textit{SemAcc}_i = \frac{1}{Os_i} \sum_{i=1}^{Os_i} \mathbb{I}\left(\cos\left(f(\hat{y}_i), f(y_i)\right) \geq \tau\right).
\end{equation}

The final overall accuracy is a weighted average: 
\begin{equation}
\textit{OverallAcc} =\frac{1}{N} \sum_{i=1}^N (\frac{Oe_i}{O_i} \cdot \textit{ExaAcc}_i + \frac{Os_i}{O_i} \cdot \textit{SemAcc}_i).
\end{equation}

\subsection{Task 2: LLM Legal Reasoning}

\subsubsection{Experimental Design}
\label{sec: task2 input modes}

This task aims to systematically evaluate the quality of multi-step reasoning processes in complex legal decision-making by LLMs. To examine the impact of reasoning CoT prompts on model behavior, we construct three input modes (see Listing \ref{Input_Modes}): 

\textbf{(1) Standard Input (Std):} Only basic role setting, task introduction, and case description are provided.

\textbf{(2) Human-Designed CoT Input (H-CoT):} Building upon standard input, we integrate authentic legal reasoning paths grounded in real-world judicial practice. These paths—distilled by legal experts from empirical case handling—systematically guide models to generate outputs through human-aligned reasoning patterns.

\textbf{(3) Self-Initiated CoT Input (S-CoT):} 
Building upon standard input, we integrate self-initiated legal reasoning paths tailored to the model. These paths, constructed by the model based on its intrinsic capabilities, are designed to unlock its latent reasoning and generation capabilities.

\begin{listingbox}[label={Input_Modes}]{Input Modes}
\textbf{Role Setting.}~You are a seasoned legal expert, proficient in legal reasoning and highly familiar with the legal regulations governing insider trading cases.

\vspace{4pt}
\textbf{Task Introduction.}~Please carefully review the case description below. Conduct a comprehensive legal analysis of the parties’ actions in conjunction with applicable statutes and evidentiary materials, culminating in a determination of their legal liability. Present a complete chain of reasoning with final conclusions.

\vspace{4pt}
\textbf{[Human-Designed CoT].}~Please conduct step-by-step legal analysis following this reasoning chain: \emph{Insider Information Formation → Information Awareness → Trading Behavior → Illegal Gains → Legal Application → Penalty Outcome.}

\vspace{4pt}
\textbf{[Self-Initiated CoT].}~Please autonomously construct a legal reasoning path based on the case facts—determine the most logically sound sequence and judgment structure. Then conduct step-by-step analysis strictly adhering to your proposed reasoning framework.

\vspace{4pt}
\textbf{Case Description.}~\ldots
\end{listingbox}

This comparative setup allows us to evaluate models' capabilities in multi-step legal reasoning, and to analyze CoT's effectiveness in enhancing reasoning quality.
 
\subsubsection{Evaluation Metrics}
To evaluate the quality of the reasoning process, We focus on two key dimensions: (1) the alignment between the reasoning traces of the model and those of the legal experts, and (2) the logical coherence of the reasoning process itself. Accordingly, we adopt a two-tier evaluation framework comprising: 

\textbf{(1) IRAC Recall}, which measures the alignment between LLM-generated reasoning traces and the expert-defined \textit{IRAC-Traces}. Let $N$ be the total number of samples. For the $i$-th sample, let $R_i$ denote the number of reasoning fields in the expert-defined reference trace, and let $O_i$ denote the number of fields correctly matched by the model. The matched fields $O_i$ consist of two types: \textbf{Exa-fields} and \textbf{Sem-fields}. The detailed matching criteria are described in Task1. IRAC Recall is defined as:

\begin{equation}
\textit{IRAC Recall} = \frac{1}{N} \sum_{i=1}^N \frac{O_i}{R_i}
\end{equation}.

\textbf{(2) LLM Score}, which measures the logical coherence of LLM-generated reasoning traces.
Following the \textit{LLM-as-a-Judge} paradigm~\citep{zheng2023judging}, we use a high-performing model (e.g., DeepSeek-R1~\cite{deepseekr1}) to automatically evaluate each generated response according to a three-level rubric (A/B/C):

% \begin{itemize}[nosep]
\begin{itemize}
  \item \textbf{A:} Logically complete and well-structured; covers most key legal elements with rigorous reasoning;
  \item \textbf{B:} Generally reasonable but contains missing steps, unclear expressions, or incomplete legal analysis;
  \item \textbf{C:} Disorganized logic, lack of justification, or incorrect application of legal provisions.
\end{itemize}

The LLM score is computed as:

\begin{equation}
\textit{LLM Score} = \frac{1 \cdot  N_{\text{A}} + 0.5 \cdot  N_{\text{B}} + 0 \cdot  N_{\text{C}}}{N},
\end{equation}
where $N$ denotes the total number of evaluated samples.

\begin{table}[t]
\centering
\begin{tabular}{lcc}
\toprule
\textbf{Model} & \textbf{FCR} & \textbf{Overall Acc} \\
\midrule
Llama3-8B-Chinese & \textbf{76.37} & 57.43 \\
GLM4-9B & 68.85 & 67.01 \\
Qwen2.5-7B & 58.12 & 57.69 \\
Qwen2.5-72B & 57.26 & 58.69 \\
Llama3.3-70B-Instruct & 57.19 & 64.72 \\
DeepSeek-V3 & 69.65 & \textbf{77.43} \\
GPT-4 & 73.28 & 77.23 \\
GPT-4o & 71.33 & 72.43 \\
\midrule
AIE-51-8-law & 29.67 & 44.28 \\
Cleverlaw & \textbf{78.43} & \textbf{61.79} \\
Law-Justice & 3.87 & 39.32 \\
Lawyer-LLM & 70.13 & 57.49 \\
\midrule
DeepSeek-R1-distill-Qwen-1.5b & 32.10 & 48.60 \\
DeepSeek-R1-distill-Qwen-7b & \textbf{75.67} & 61.71 \\
QwQ-32b & 71.51 & 65.74 \\
DeepSeek-R1-distill-Llama-70B & 52.72 & 63.17 \\
DeepSeek-R1 & 70.24 & 70.72 \\
o1-mini & 64.99 & \textbf{76.45} \\
o3-mini & 66.19 & 74.39 \\
\bottomrule
\end{tabular}
\caption{Comparative performance of LLMs on the automatic annotation task. The first block of rows corresponds to General LLMs, the second block corresponds to Legal LLMs, and the third block corresponds to Reasoning LLMs.}
\label{annotion_task}
\end{table}

\section{Results and Analysis}
We show the potential of MSLR as a useful benchmark for LLMs across multiple dimensions of interest. Here are some key findings:

\textbf{- LLM-based automated annotation underperforms on tasks involving highly complex and nuanced reasoning trajectories.} As shown in Table \ref{annotion_task}, LLMs—including large-scale models like DeepSeek and GPT series—fail to exceed 80\% in both FCR and Overall Accuracy on the MSLR dataset. This indicates that LLM automated annotation falls short of achieving quality comparable to human annotators in terms of both comprehensiveness and accuracy. These results align with prior findings \cite{ChatGPTJack2023, ChatGPTLabel2024}, which highlight that LLMs still exhibit limitations in zero-shot performance on more challenging and practical tasks compared to human experts. Thus, our PPP annotation pipeline is not only necessary but also validates a promising paradigm: Human-LLM collaborative annotation \cite{CoAnnotating2023, HumanLLMCollaborativeAnnotation2024}, where LLMs accelerate the process while preserving the high quality of human-reviewed annotations.

\begin{figure}[htbp]
\centering
\includegraphics[width=0.6\columnwidth]{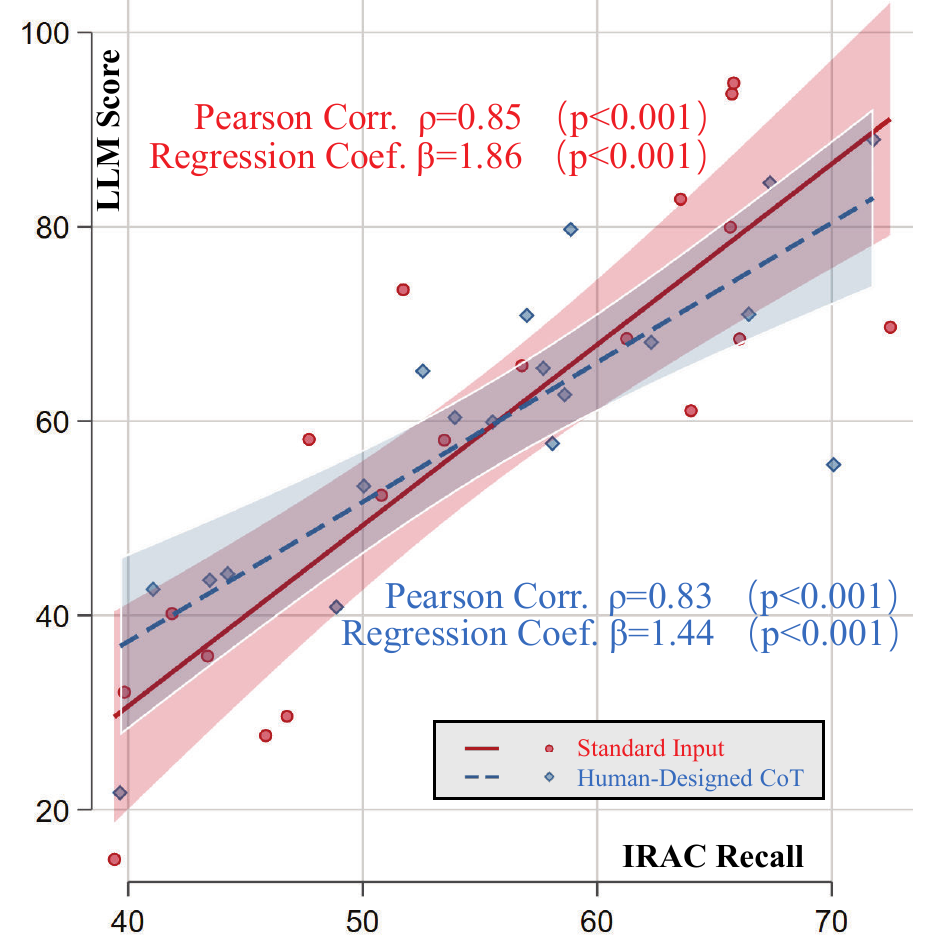} 
\caption{Pearson correlation analysis and linear fit between IRAC Recall and LLM Score.}
\label{corr}
\end{figure}

\textbf{- The LLM-as-a-Judge paradigm provides a scalable method for evaluating the quality of legal reasoning, with validity confirmed through dual verification.} First, Figure \ref{corr} shows a statistically significant positive correlation between IRAC Recall and LLM Score, indicating strong metric alignment. Second, legal experts independently assessed 380 randomly sampled reasoning outputs (20 outputs from each of 19 models). A comparison between these human ratings and automated ratings from DeepSeek-R1 reveals an 87.96\% agreement rate in grade classification, demonstrating high human-machine consistency. These results confirm that model-generated reasoning steps can be evaluated both consistently and accurately via LLM-as-a-judge, closely aligning with human expert assessments. Our work demonstrates that general-domain LLMs are capable of assessing the quality of legal reasoning.

\begin{figure}[t]
\centering
\includegraphics[width=0.5\columnwidth]{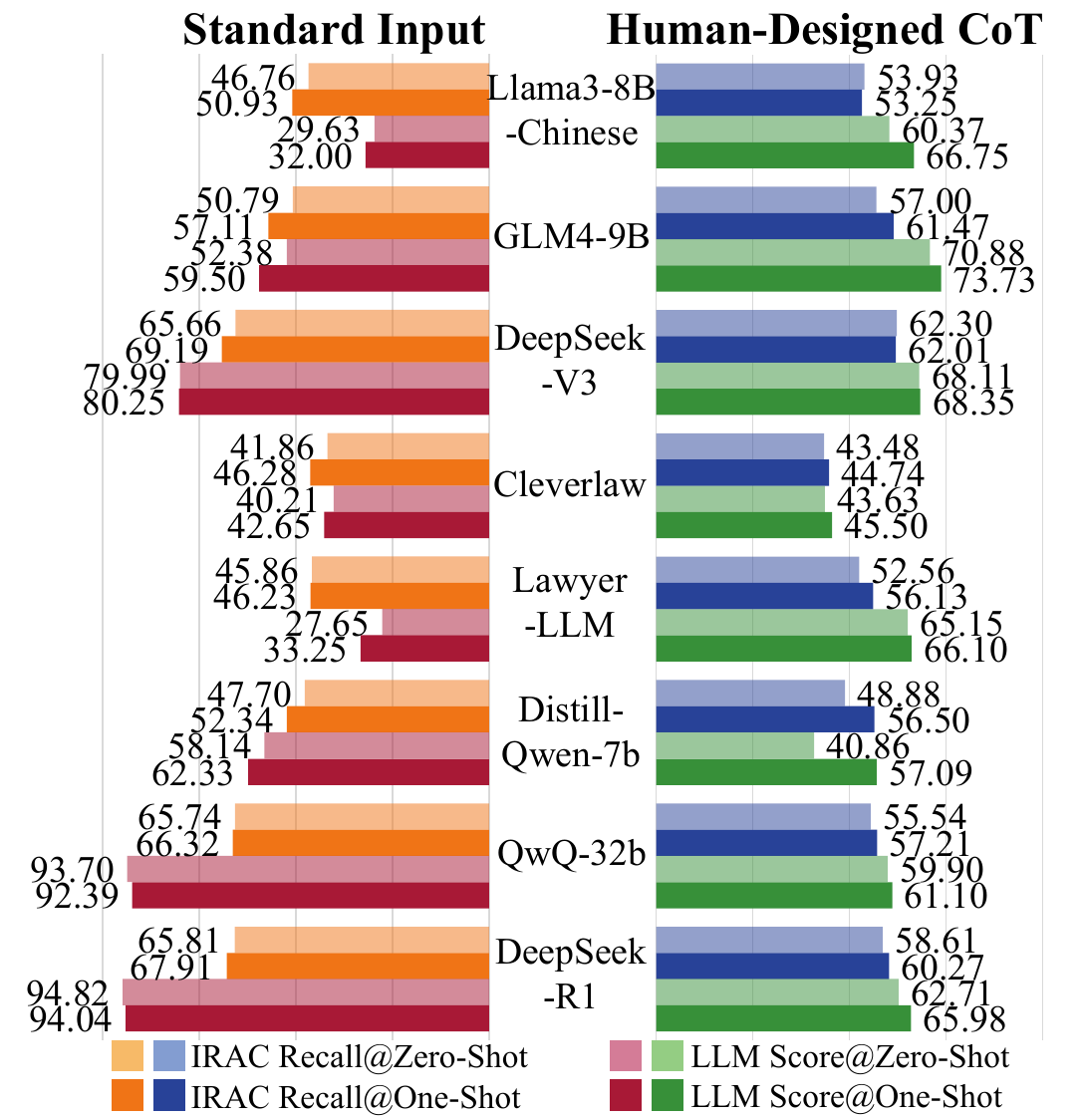} 
\caption{Zero-Shot vs. One-Shot performance on the legal reasoning task.}
\label{fig: oneshot}
\end{figure}

\textbf{- LLMs exhibit limited proficiency in multi-step legal reasoning.} As shown in Table \ref{reasoning_task}, the overall performance of all evaluated models remains suboptimal. Although some large-scale models(e.g., QwQ-32B, Deepseek-R1) achieve LLM Score exceeding 90\%, their IRAC Recall consistently falls short of 75\%. This indicates that while the generated reasoning may appear logically coherent on the surface, it often lacks alignment with the structured reasoning paths used in real-world legal decision-making. As previously analyzed, there is a significant positive correlation between IRAC Recall and LLM Score, further validating that IRAC Recall serves as a more objective and structured metric for assessing models’ true reasoning capabilities in legal tasks. 

To further explore models’ performance under few-shot settings, we conducted additional experiments. Due to the considerable length of legal documents, we adopted a one-shot setup in our experiments (see Figure \ref{fig: oneshot}). Results show that while the one-shot setting leads to a moderate improvement in both IRAC Recall and LLM Score, the overall performance gains remain limited. This finding aligns with recent findings \cite{InterpretableLongformLegal2024a, shenLawReasoningBenchmark2025, LEXam2025}, which highlight significant room for improvement in LLMs’ handling of complex legal reasoning tasks.

\begin{table*}[htbp]
\centering
\begin{tabular}{l c c c c c c}
\toprule
 \textbf{Model} &
\multicolumn{2}{c}{\textbf{Standard Input}}&
\multicolumn{2}{c}{\textbf{Human-Designed CoT}}&
\multicolumn{2}{c}{\textbf{Delta}}\\
 & \textbf{IRAC Recall}& \textbf{LLM Score }& \textbf{IRAC Recall}& \textbf{LLM Score }& \textbf{IRAC Recall}&\textbf{LLM Score }\\
\midrule
 Llama3-8B-Chinese & 46.76 & 29.63 & 53.93 & 60.37 & \textbf{7.17} & 30.74 \\
  GLM4-9B & 50.79 & 52.38 & 57.00 & 70.88 & 6.21 & 18.50 \\
  Qwen2.5-7B & 56.78 & 65.73 & 57.70 & 65.44 & 0.92 & -0.29 \\
  Qwen2.5-72B & 61.25 & 68.50 & 58.09 & 57.70 & -3.16 & -10.80 \\
  Llama3.3-70B-Instruct & 53.47 & 58.06 & 58.87 & 79.73 & 5.40 & \textbf{21.67} \\
  DeepSeek-V3 & 65.66 & \textbf{79.99} & 62.30 & 68.11 & \textbf{-3.36} & \textbf{-11.88} \\
  GPT-4 & 63.99 & 61.09 & 66.46 & 70.99 & 2.47 & 9.90 \\
  GPT-4o & \textbf{66.06} & 68.47 & \textbf{71.78} & \textbf{88.98} & 5.72 & 20.51 \\
\midrule
 AIE-51-8-law & 39.41 & 14.90 & 39.65 & 21.74 & 0.24 & 6.84 \\
  Cleverlaw & 41.86 & \textbf{40.21} & 43.48 & 43.63 & 1.62 & 3.42 \\
  Law-Justice & 39.83 & 32.11 & 41.06 & 42.69 & 1.23 & 10.58 \\
  Lawyer-LLM & \textbf{45.86} & 27.65 & \textbf{52.56} & \textbf{65.15} & \textbf{6.70} & \textbf{37.50} \\
\midrule
 Distill-Qwen-1.5b & 43.38 & 35.82 & 44.24 & 44.28 & 0.86 & \textbf{8.46} \\
  Distill-Qwen-7b & 47.70 & 58.14 & 48.88 & 40.86 & 1.18 & -17.28 \\
  QwQ-32b & 65.74 & 93.70 & 55.54 & 59.90 & \textbf{-10.20} & \textbf{-33.80} \\
  Distill-Llama-70B & 51.72 & 73.54 & 50.05 & 53.31 & -1.67 & -20.23 \\
  DeepSeek-R1 & 65.81 & \textbf{94.82} & 58.61 & 62.71 & -7.20 & -32.11 \\
  o1-mini & \textbf{72.49} & 69.69 & \textbf{70.08} & 55.51 & -2.41 & -14.18 \\
  o3-mini & 63.55 & 82.87 & 67.37 & \textbf{84.52} & \textbf{3.82} & 1.65 \\
\bottomrule
\end{tabular}
\caption{Comparing Standard Input and Human-Designed CoT on the legal reasoning task.}
\label{reasoning_task}
\end{table*}

\textbf{- The impact of identical CoT prompts varies across LLMs, yielding both positive and negative effects.} As Table \ref{reasoning_task} indicates, Human-Designed CoT prompts consistently enhance IRAC Recall and LLM Score for all legal models and most general models. This demonstrates that infusing models with real-world reasoning paths improves reasoning quality and logical coherence. Conversely, for specific general models (e.g., Qwen2.5-72B, DeepSeek-V3) and most reasoning models suffer systematic performance degradation with Human-Designed CoT. For instance, the QwQ-32B model experiences an absolute decrease of 33.8\% in LLM Score and a 10.20\% drop in IRAC Recall. Case analyses by legal experts have uncovered a manifestation of cognitive interference from structural guidance. Specifically, while Human-Designed CoT formally enhances the structured coherence of reasoning traces, it forces reasoning models with intrinsic inference mechanisms to conform to external logical frameworks. This deviation from their native reasoning pathways ultimately leads to inappropriate application of legal provisions and fabricated analytical content.  These findings challenge the prevailing assumption of universal CoT benefits and underscore the open question of when CoT systematically reduces performance \cite{CoTNotCoT2025, MindYourStep2025}.

\begin{figure}[htbp]
\centering
\includegraphics[width=0.7\columnwidth]{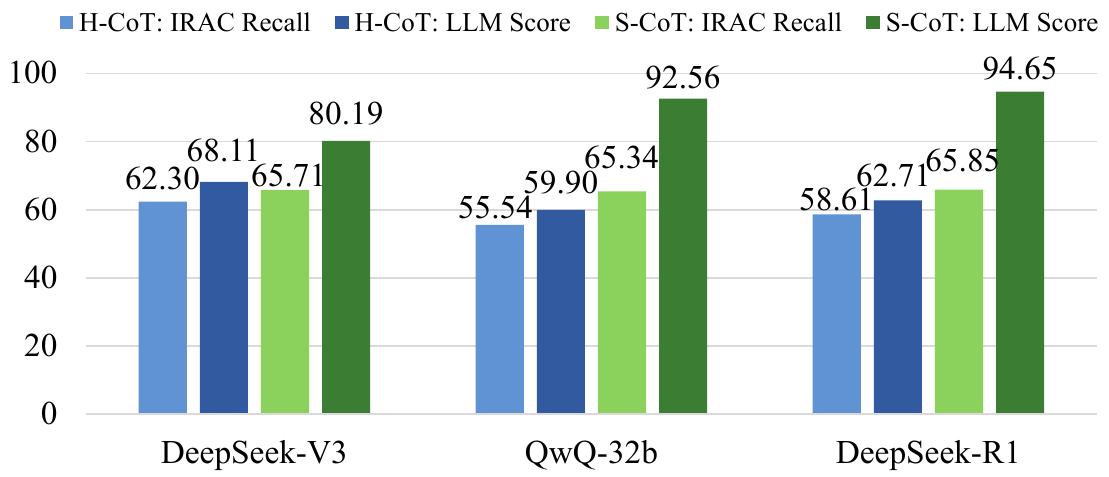} 
\caption{H-CoT vs. S-CoT on the legal reasoning task.}
\label{S-CoT}
\end{figure}

\begin{figure}[t]
\centering
\includegraphics[width=0.6\columnwidth]{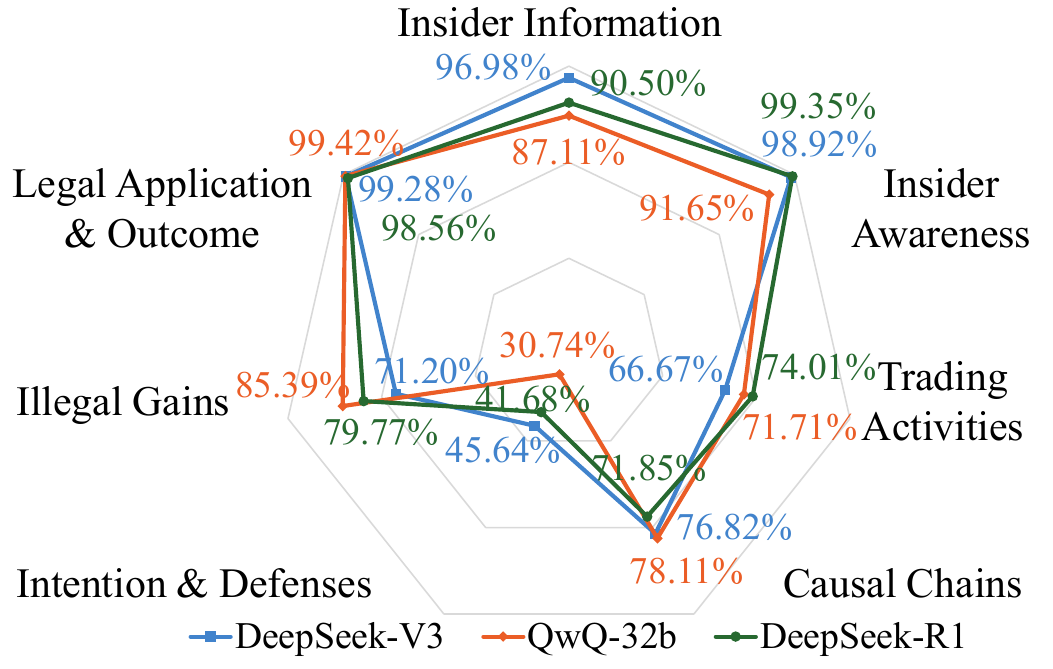} 
\caption{Typology and coverage of Self-Initiated CoT reasoning traces in three mainstream LLMs.}
\label{SCOT-output}
\end{figure}

\textbf{- To resolve the inherent tension between prompt rigidity and model capability, the Self-Initiated CoT prompt strategy emerges as a promising paradigm.} Inspired by Wang et al. \cite{SelfConsistencyCot2023} and Jin et al. \cite{SelfHarmonized2025}, we implement Self-Initiated CoT that guide models to autonomously generate reasoning steps before legal analysis. As Figure \ref{S-CoT} demonstrates, these prompts outperform Human-Designed CoT in enhancing reasoning quality and logical coherence. Legal experts validated this approach by categorizing output traces, revealing through regex matching and information extraction that model-generated paths consistently cover seven core legal elements (see Figure \ref{SCOT-output}). Results indicate that while autonomously constructed paths exhibit diversity, they maintain comprehensiveness. We attribute this to: (1) The ``Hydra Effect'' \cite{HydraEffect2023} - where blocked reasoning routes activate alternative pathways, mirroring real-world legal problem-solving; (2) Transformers' parallel information processing across distributed components, contrasting with sequential CoT constraints \cite{HowThinkStepbystep2024}. This phenomenon underscores that model-task aligned prompting is critical for achieving high-quality reasoning.

\section{Conclusion and Future Work}

In this work, we introduce \textbf{MSLR}, a specialized benchmark for multi-step legal reasoning tasks designed to evaluate both reasoning process quality and logical coherence. Comprising nearly 1,400 legal decisions requiring an average of over 40 detailed reasoning steps each, MSLR addresses critical gaps in assessing fine-grained legal reasoning. To enhance annotation efficiency and mitigate semantic understanding biases in LLM-generated annotations, we further design a scalable Human-LLM collaborative annotation pipeline. Systematic evaluations of various LLMs in MSLR reveal substantial limitations in the legal reasoning capabilities of current LLMs. Crucially, our analysis demonstrates that Self-Initiated CoT outperforms Human-Designed CoT prompts. Establishing model-task aligned prompting strategy emerges as a promising paradigm to enable LLMs to achieve high-quality reasoning and stable outputs.

\section{Acknowledgments}
This work was supported by the National Science Fund for Excellent Young Scholars (Overseas) under grant No.\ KZ37117501, the National Natural Science Foundation of China (No. 62306024), the Key Program of the National Social Science Fund of China (No. 23AFX002), the Fundamental Research Funds for the Central Universities, and the Center for Empirical Legal Studies of Shanghai Jiao Tong University.

\bibliography{references}

\clearpage
\appendix
\section{Data and Code}

The data and code for this work are available in the folder ``\texttt{MSLR-Multi-Step-Reasoning-Trace}''.

\begin{table*}[htbp]
\centering
\begin{tabular}{llll}\toprule
 \textbf{File from Lawyer LLaMA}& \textbf{ROUGE-1 F1}& \textbf{ROUGE-2 F1 }& \textbf{ROUGE-L F1}\\
\midrule
 alpaca\_gpt4\_data\_zh.json
& 0.0303& 0.0033& 0.0247
\\
  fakao\_gpt4.json
& 0.0171& 0.0008& 0.0155
\\
  judical\_examination\_v2.json
& 0.0083& 0.0007& 0.0071
\\
  legal\_counsel\_multi\_turn\_with\_article\_v2.json
& 0.0032& 0.0002& 0.0029
\\
  legal\_counsel\_v2.json
& 0.0168& 0.0026& 0.0144
\\
  zixun\_gpt4.json & 0.0017& 0.0001& 0.0014\\ \bottomrule
\end{tabular}
\caption{Document similarity between MSLR and Lawyer LLaMA.}
\label{rouge_result}
\end{table*}

\section{The Novelty of MSLR}

To validate the novelty of the MSLR dataset, we compute ROUGE-1/2/L $F_1$ scores between documents in MSLR and the comprehensive Chinese legal corpus used during the training of Lawyer LLaMA. As shown in Table \ref{rouge_result}, the results show near-random similarity levels (all \(\ F_1 < 0.03 \)), significantly below conventional similarity thresholds. This indicates that MSLR was not included in the pretraining corpus, further establishing its reliability as a contamination-free evaluation benchmark.

To ensure scientific integrity and impartiality, MSLR incorporates systematic anti-leakage protocols across four dimensions: (1) Transparent data sourcing. All legal documents are collected from authoritative platforms (e.g., the China Securities Regulatory Commission (CSRC), the Supreme People's Procuratorate (SPP), and the Supreme People's Court (SPC)) with preserved timestamps and source URLs, ensuring traceability and legal compliance. (2) Inference-centric design. The tasks in MSLR are inherently reasoning-intensive, making it impractical to achieve high scores through superficial text memorization or reproduction. (3) Rigorous evaluation. Large language models (LLMs) are strictly evaluated under zero-shot or one-shot settings without any training or fine-tuning on MSLR dataset, guaranteeing objective assessment. (4) Read-only API usage. Read-only interfaces prevent feedback loop contamination. During model invocation, only read-only APIs are used, and evaluation outputs are never fed back into model training or corpus updates, eliminating feedback-based contamination entirely. Taken together, the MSLR architecture mitigates leakage risks while establishing its scientific validity, impartiality, and challenge level for legal AI evaluation.

\section{LLM Automatic Annotation Detail}

\subsection{Structured Data Annotation Template}
Based on these IRAC-Traces, we develop a structured data annotation template to systematically record the reasoning process of each legal decision. As shown in the Listing \ref{Annotation_Template}, the template includes metadata of legal documents, key structured elements essential for legal reasoning and  decision-making, as well as the corresponding legal provisions and factual case descriptions. Each case instance is composed of multiple hierarchical fields, with a total of over \textbf{40} defined attributes, ensuring both completeness of representation and traceability of the reasoning process. 

\subsection{Standardized Prompts for Task 1}
\label{app:task1 prompt}

To evaluate the annotation capabilities of LLMs in the legal domain (i.e., Task 1), we transform annotation tasks into structured information extraction tasks compatible with LLMs. These tasks encompass core NLP sub-tasks including named entity recognition, text classification, and summarization. We design a unified framework for structured information extraction, where prompts are constructed under a zero-shot setting. As shown in the Listing \ref{Annotation_Prompt}, this framework employs a standardized input template to guide LLMs in extracting key legal information from raw case texts. The output is required to follow the predefined JSON schema (see Listing \ref{Annotation_Template}), facilitating downstream evaluation and processing. It is important to note that both the inputs and outputs are in Chinese; therefore, the actual prompt is written in Chinese, and the English version presented here is a direct translation for illustration purposes.
\begin{listingbox}[label={Annotation_Prompt}]{Standardized Prompts for Task 1}

\textbf{Role Setting:} \\
You are a large language model specialized in legal document analysis and information extraction, particularly skilled at identifying legal facts related to insider trading. Please read the following case description and extract the specified fields, outputting a standardized structured JSON. \\[6pt]
\textbf{Task Introduction:} \\
(1) Please strictly extract only the fields listed under the ``Field Extraction Scope'' below; \\
(2) If a field is not mentioned in the Case Description, leave it as an empty string ``''; \\
(3) The output must be a valid JSON object, and the field names and nested structure must exactly follow the ``Output Format Template''; \\
(4) Do not generate any explanations or comments—output only the JSON data. \\[6pt]
\textbf{Case Description:} \\
... \\[6pt]
\textbf{Field Extraction Scope:} \\
... \\[6pt]
\textbf{Output Format Template:} \\
... (see Listing \ref{Annotation_Template} ``Structured Data Annotation Template'')

\end{listingbox}

\subsection{Prompt Optimization in the Precision-Tuning Stage}
\label{app:generate-prompt}

In the precision-tuning stage, we evaluate different LLMs' performance against gold-standard examples to identify the most cost-effective model. Experimental results demonstrate DeepSeek-V3's optimal cost-effectiveness ratio, establishing it as our primary annotation assistant. Building on the optimal model, we further refine prompts for a subset of fields that are semantically dependent, ambiguous, or irregularly formatted. By optimizing prompts and conducting independent extraction trials, we significantly enhance the accuracy of information extraction for these challenging fields. The following presents key optimization strategies for selected fields, along with illustrative prompt examples.

\subsubsection{Content of Insider Information}

This field is often composed of multiple factual descriptions and lacks clearly defined boundaries. As shown in Listing \ref{Content_of_Insider_Information}, we apply a summarization-style prompt that directs the model to distill the ``core insider information'' into 1–2 legally framed sentences.

\begin{listingbox}[label={Content_of_Insider_Information}]{Prompt for \textit{Content of Insider Information} Field}

Please summarize the core facts of the insider information in 1–2 sentences using formal legal language. For example: ``The major asset restructuring information of the company was leaked before public disclosure ...''

\end{listingbox}

\subsubsection{Type of Party Involved}

This field requires complex classification based on Article 51 of the \textit{Securities Law} and Article 6 of the \textit{CSRC's Guidelines for Identifying Insider Trading}. The classification is multi-label by nature. As shown in Listing \ref{Type_of_Party_Involved}, we design the prompt to reference relevant legal clauses and support multiple-choice outputs from a provided candidate list.

\begin{listingbox}[label={Type_of_Party_Involved}]{Prompt for \textit{Type of Party Involved} Field}

According to Article 51 of the \textit{Securities Law} and Article 6 of the \textit{CSRC's Guidelines for Identifying Insider Trading}, determine the category of insider the involved party belongs to. Multiple selections are allowed from the following candidate list:

``51(1) Issuer and its directors, supervisors, senior executives'';

...

``Guideline 6(2-1) Issuer or listed company'';

``Guideline 6(2-2) Other companies controlled by controlling shareholders of the issuer'';

...

\end{listingbox}

\subsubsection{Legal Provisions Applied}

This field often suffers from variability in expression, with frequent omissions of article numbers or formal statute names. To mitigate this, as shown in Listing \ref{Legal_Provisions_Applied}, we explicitly instruct the model to extract valid legal references, including both the statute title and the article number. The scope is limited to formal legal documents such as the \textit{Securities Law} and \textit{Criminal Law}.

\begin{listingbox}[label={Legal_Provisions_Applied}]{Prompt for \textit{Legal Provisions Applied} Field}

Please extract the legally effective article names and numbers from the following legal description. Only consider provisions from formal legal documents such as the \textit{Securities Law} and \textit{Criminal Law}. For example: ``Article 73 of the Securities Law''; ...

\end{listingbox}

\subsubsection{Summary of Objections or Defenses}

This field is prone to confusion with the adjudicator's opinions. To avoid this, as shown in Listing \ref{Summary_of_Objections_or_Defenses}, the prompt includes a perspective instruction emphasizing that only content asserted by the party should be extracted.

\begin{listingbox}[label={Summary_of_Objections_or_Defenses}]{Prompt for \textit{Summary of Objections or Defenses} Field}

Only extract the objections or defenses claimed by the involved party, excluding comments from the court or regulatory authorities. Begin the summary with ``The party argued that ...''.

\end{listingbox}

\section{LLM-as-a-Judge Evaluation Template}
\label{app:evaluation-prompt}
We adopt an evaluation setup based on the \textbf{LLM-as-a-Judge} paradigm to measure the quality and logical coherence of LLM-generated reasoning traces. We select the flagship model \textbf{DeepSeek-V3} as the evaluation model and design a standardized input format. The complete evaluation prompt is shown in the Listing \ref{evaluation_prompt}.

\begin{listingbox}[label={Annotation_Template}]{Structured Data Annotation Template}
\textbf{Case Metadata}: 
\begin{lstlisting}
"Case Index": "...",
"Document Type": "...",
"Final Penalty Outcome": "...",
"Penalty Year": "...",
"Penalty Date": "...",
"Penalizing Authority": "...",
"Case Reference Number": "...",
"Title of the Decision Document": "..."
\end{lstlisting} 

\textbf{Insider Information Determination}: 
\begin{lstlisting}
"Name of the Traded Stock": "...",
"Content of Insider Information": "...",
"Legal Basis for Insider Information Determination": "...",
"Type of Insider Information": "...",
"Date of Insider Information Formation": "...",
"Description of Events on Formation Date": "...",
"Date of Public Disclosure": "...",
"Description of Events on Disclosure Date": "..."
\end{lstlisting}

\textbf{Basic Information of the Party}: 
\begin{lstlisting}
"Name": "...",
"Gender": "...",
"Year of Birth": "...",
"Education Level": "...",
"Position at the Time of the Incident": "..."
\end{lstlisting}

\textbf{Insider Trading Determination for the Party}: 
\begin{lstlisting}
"Role of the Party": "...",
"Type of Party Involved": "...",
"Date the Party Became Aware of Insider Information": "...",
"Manner in Which Insider Information Was Obtained": "...",
"Classification of Access Method": "...",
"Type of Insider Trading Behavior": "...",
"Buy/Sell Action": "...",
"Date(s) of Trade Execution": "...",
"Trade Amount (in RMB)": "...",
"Earliest Trade Date": "...",
"Latest Trade Date": "...",
"Benchmark Value on Key Date": "...",
"Illicit Gains (in RMB)": "..."
\end{lstlisting}

\textbf{Penalty Details}: 
\begin{lstlisting}
"Legal Provisions Applied": "...",
"Amount of Illicit Gains Confiscated (in RMB)": "...",
"Multiplier of Fine": "...",
"Amount of Fine (in RMB)": "..."
\end{lstlisting}

\textbf{Arguments Raised by the Party}: 
\begin{lstlisting}
"Summary of Objections or Defenses": "..."
\end{lstlisting}

\textbf{Legal Decision Text}: 
\begin{lstlisting}
"Original Text of the Legal Document": "...",
"Structured Factual Description": "...",
"Summary of Legal Reasoning": "...",
"Judgment or Penalty Decision": "..."
\end{lstlisting}
\end{listingbox}

\begin{listingbox}[label={evaluation_prompt}]{LLM-as-a-Judge Evaluation Prompt}

\textbf{Role Description}

You are a legal expert reviewer tasked with evaluating and comparing the legal analysis results generated by the same AI model under three different modes. The model is expected to perform a complete analysis based on the following legal reasoning chain:

\textit{Insider Information Formation $\rightarrow$ Information Awareness $\rightarrow$ Trading Behavior $\rightarrow$ Illegal Gains $\rightarrow$ Legal Application and Judgment Type $\rightarrow$ Penalty Outcome}
\\

\textbf{Scoring Criteria}

Please rate the reasoning quality of the outputs according to the following criteria:

\begin{itemize}
  \item \textbf{A (Excellent)}: The reasoning fully covers all six stages above, with clear and rigorous logic, correct application of law, and a reasonable conclusion.
  \item \textbf{B (Fair)}: The reasoning is generally reasonable but contains omissions or ambiguities in certain stages, affecting the rigor of legal judgment.
  \item \textbf{C (Poor)}: The reasoning is severely lacking, logically inconsistent, or legally incorrect, making it insufficient to support a valid judgment.\\
\end{itemize} 

\textbf{Output Format}

Please return the three scores strictly in the following format (A, B, or C):

\texttt{Standard Score: <score>} \\
\texttt{Human-Designed CoT Score: <score>} \\
\texttt{Self-Initiated CoT Score: <score>} \\

\textbf{Input Format}

\noindent\textit{- Standard Output:} \\
\texttt{\{predicted\_answers[0]\}} \\
\noindent\textit{- Human-Designed CoT Output:} \\
\texttt{\{predicted\_answers[1]\}} \\
\noindent\textit{- Self-Initiated CoT Output:} \\
\texttt{\{predicted\_answers[2]\}} \\
\noindent\textit{- Reference Answer:} \\
\texttt{\{Summary of Legal Reasoning\} + \{Judgement or Penally Decision\}}

\end{listingbox}

% \newpage
\section{Field Categorization Details}
\label{app:field-categories}
To enhance the precision and interpretability of evaluations for both automatic annotation and legal reasoning tasks, We classify fields into two categories: \textbf{Exa-Fields} requiring exact matching (e.g.,\textit{dates}, \textit{amounts}) and \textbf{Sem-Fields} requiring semantic matching (e.g., \textit{event descriptions}, \textit{information summaries}). This classification facilitates a more fine-grained and legally grounded evaluation strategy, better aligning with the inherent structural characteristics of legal data. The detailed categorization is presented in Table~\ref{field-categories}.

\begin{table}[H]
\centering
\renewcommand{\arraystretch}{1.2}
\begin{tabular}{l|l|l}\toprule
 \textbf{Field Type}& \textbf{Section}& \textbf{Field Name}\\
\midrule
 \multirow{18}{*}{Exa-Fields}& & Name of the Traded Stock\\
 & Insider Information Determination & Date of Insider Information Formation\\
 & & Date of Public Disclosure\\
  \cline{2-3}
 & & Name\\
 & Basic Information of the Party & Gender\\
 & & Year of Birth\\
  & & Position at the Time of the Incident\\
   \cline{2-3}
  & & Date of Awareness\\
 & &Buy/Sell Action\\
 & &Date(s) of Trade Execution\\
 & Insider Trading Determination for the Party &Trade Amount (RMB)\\
  & & Earliest Trade Date\\
 & &Latest Trade Date\\
 & &Benchmark Value on Key Date\\
 & &Illicit Gains (RMB)\\
   \cline{2-3}
 & Penalty Outcome Details&Amount of Illicit Gains Confiscated (RMB)\\
 & &Multiplier of Fine \\
 & &Amount of Fine (RMB)\\
 \midrule
 \multirow{11}{*}{Sem-Fields}& &Content of Insider Information\\
 & &Legal Basis for Insider Information Determination\\
 & Insider Information Determination &Type of Insider Information\\
 & &Description of Event  on Formation Date\\
 & &Description of Event on Disclosure Date\\
 \cline{2-3}
 & &Role of the Party\\
 & &Type of Party Involved\\
 & Insider Trading Determination for the Party &Manner in Which Insider Information Was Obtained\\
 & &Classification of Access Method\\
 & &Type of Insider Trading Behavior\\
 \cline{2-3}
 & Penalty Outcome Details&Legal Provisions Applied\\
\midrule
\end{tabular}
\caption{Field categorization for evaluation.}
\label{field-categories}
\end{table}

\newpage
\section{Supplementary Experiments for Task 1}

We conducted supplementary one-shot experiments on the LLM automatic annotation task (Task 1). As shown in Table \ref{oneshot_task1}, under the one-shot setting, most models demonstrated moderate improvements in Field Completeness Rate (FCR) compared to zero-shot, indicating that demonstrative example partially enhance field extraction comprehensiveness. However, Overall Accuracy (Overall Acc) shows negligible improvement, with some models exhibiting performance degradation. This reveals that while one-shot prompting expands output coverage, it fails to significantly enhance extraction precision. These findings demonstrate that minimal examples cannot overcome LLMs' limitations in parsing complex legal semantics. Consequently, our established PPP annotation pipeline—leveraging Human-LLM collaboration to accelerate processing while maintaining human-curated quality—proves both necessary and promising.

\begin{table}[H]
\centering
\small
\begin{tabular}{l l cc cc}
\toprule
\textbf{Category} & \textbf{Model} & \multicolumn{2}{c}{\textbf{Zero-shot}} & \multicolumn{2}{c}{\textbf{One-shot}} \\
 &  & \textbf{FCR} & \textbf{Overall Acc} & \textbf{FCR} & \textbf{Overall Acc} \\
\midrule
\multirow{3}{*}{General LLMs}
 & Llama3-8B & 76.37 & 57.43 & 76.40 & 58.50 \\
 & GLM4-9B & 68.85 & 67.01 & 61.37 & 57.17 \\
 & DeepSeek-V3 & 69.65 & 77.43 & 73.07 & 75.11 \\
\midrule
\multirow{2}{*}{Legal LLMs}
 & Cleverlaw & 78.43 & 61.79 & 64.76 & 55.73 \\
 & Lawyer-LLM & 70.13 & 57.49 & 83.77 & 58.91 \\
\midrule
\multirow{3}{*}{Reasoning LLMs}
 & Distill-Qwen-7b & 75.67 & 61.71 & 80.11 & 57.81 \\
 & QwQ-32b & 71.51 & 65.74 & 75.54 & 63.90 \\
 & DeepSeek-R1 & 70.24 & 70.72 & 77.61 & 71.71 \\
\bottomrule
\end{tabular}
\caption{Zero-shot vs. One-shot performance on the automatic annotation task.}
\label{oneshot_task1}
\end{table}

\section{Case Study for Task 2}

\begin{figure}[thpb]
\centering
\includegraphics[width=0.9\textwidth]{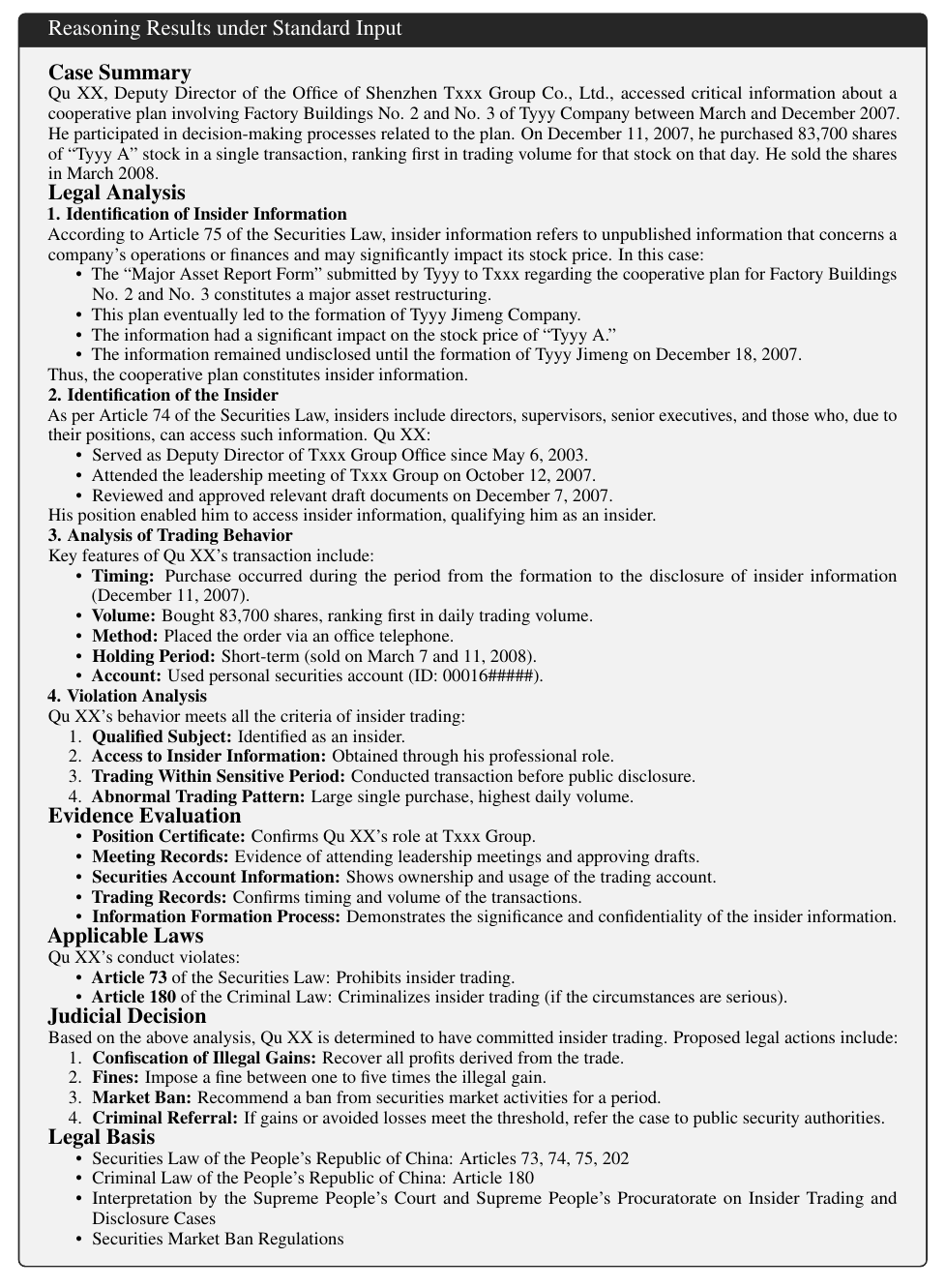}% Reduce the figure size so that it is slightly narrower than the column.
\caption{Reasoning results under Standard input.
}
\label{std_output}
\end{figure}

\begin{figure}[thpb]
\centering
\includegraphics[width=0.89\textwidth]{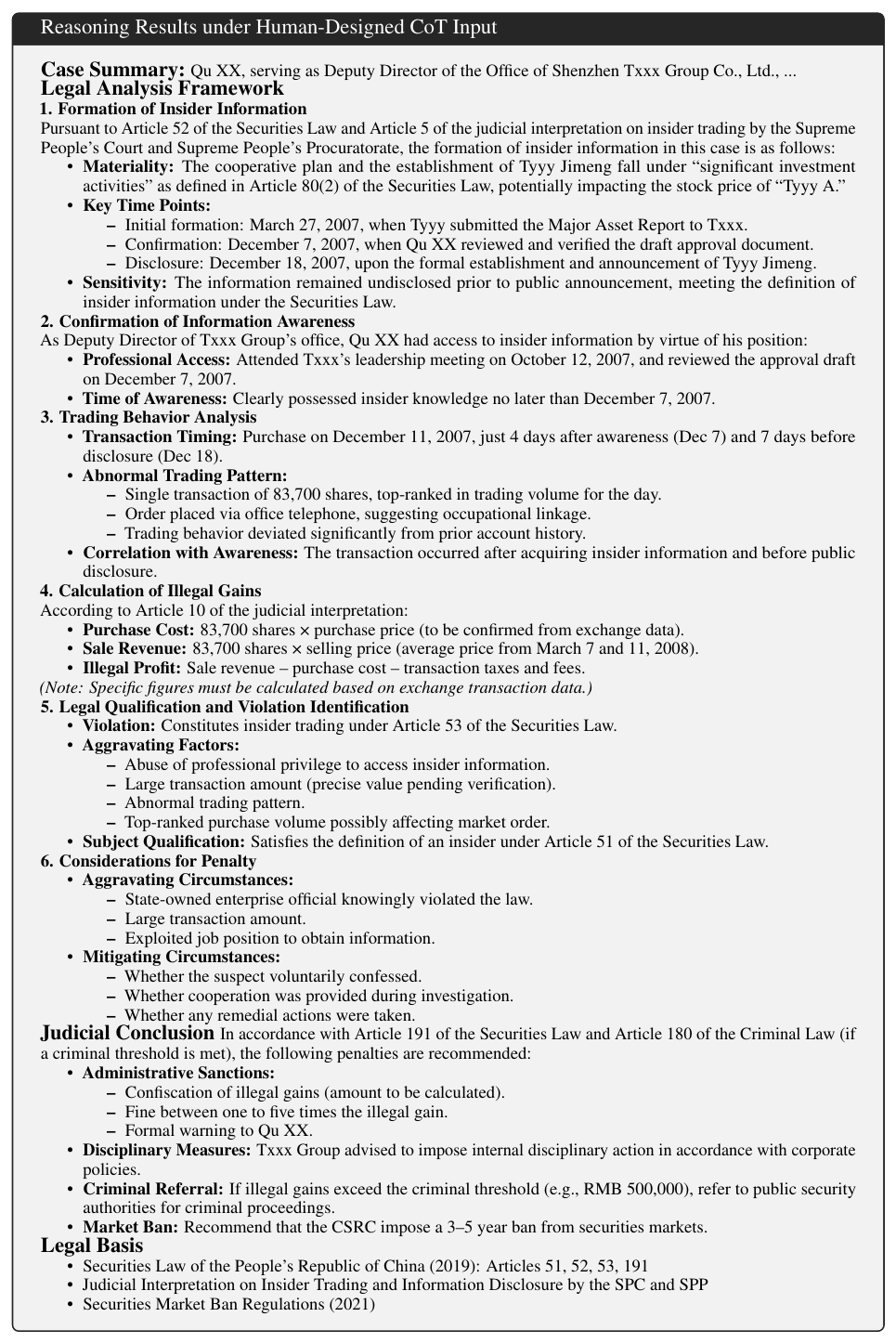}% Reduce the figure size so that it is slightly narrower than the column.
\caption{Reasoning results under Human-Designed CoT input.}
\label{hcot_output}
\end{figure}

In this case study, we present and analyze the reasoning processes generated by the \textbf{DeepSeek-R1} model when processing the same case description under two prompt modes: Standard Prompt and Human-Designed CoT Prompt. Their generated reasoning processes are respectively shown in Figure \ref{std_output} and Figure \ref{hcot_output}. A detailed comparison of their reasoning quality reveals that under LLM-as-a-Judge, the output from the Standard prompt mode received an A-grade rating, whereas the Human-Designed CoT prompt output was only awarded a C-grade. Legal experts' assessments (as shown in Table 5) align with the model ratings, confirming that the Standard prompt reasoning process better adheres to case facts and legal application, demonstrating significant advantages over the Human-Designed CoT prompt in terms of logical structure and legal coherence. This case study highlights a critical limitation of Human-Designed CoT: despite its intent to guide systematic reasoning, its redundant structural framework may disrupt the model’s intrinsic reasoning mechanisms, leading to logical leaps, missing elements, or incorrect application of legal provisions—all of which undermine the rationality and credibility of the final judgment.

\begin{table}[H]
\centering
\small
\begin{tabular}{|p{0.1\linewidth}|p{0.3\linewidth}|p{0.3\linewidth}|p{0.2\linewidth}|}
\hline
\textbf{Analysis Dimension} & \textbf{Standard Output} & \textbf{Human-Designed Output}& \textbf{Evaluation Conclusion} \\
\hline
Case Fact Recognition & Accurate and complete, covering the full process of insider information formation and trading & Omits the early stage of insider information formation, which does not affect substantive determination & Roughly equivalent \\
\hline
Insider Information Identification & Correctly cites Article 75 of the 2005 Securities Law; applies the historical law accurately & Incorrectly cites the 2019 version and Article 80 (misapplies information disclosure clauses) & Std clearly superior; CoT lacks legal clarity \\
\hline
Insider Awareness Analysis & Reasonably judged based on Article 74, considering official position and timing of awareness & Introduces a “time of awareness” path, but the logical chain is unclear and partially fabricated & Std aligns better with legal logic; CoT introduces invented reasoning points\\
\hline
Trading Behavior Analysis & Covers timing, amount, method, and account details; relatively well-structured & Introduces “habitual trading comparison” as a novel point, but lacks conclusive judgment & Both have weaknesses \\
\hline
Illegal Gains Calculation & Not provided & Provided, but based on incorrect legal citation & Both have deficiencies \\
\hline
Legal Norm Application & Comprehensive citation of historical laws and judicial interpretations; reflects administrative-criminal transition & Incorrectly cites new legal provisions; omits criminal transfer criteria & Std more accurate; CoT misuses legal references\\
\hline
 \end{tabular}
\caption{Comparison of Standard Output vs. Human-Designed CoT Output on the legal reasoning task.}
\label{tab:std-vs-cot}
\end{table}

\end{document}